\documentclass[sigconf, nonacm]{acmart}
\usepackage{multirow}
\usepackage{tablefootnote}

\AtBeginDocument{%
  }

\setcopyright{none}

\usepackage{booktabs}
\usepackage{graphicx}
\usepackage{longtable}   
\usepackage{tabularx,booktabs}
\begin{document}
\title{Reasoning or Overthinking: Evaluating Large Language Models on Financial Sentiment Analysis
}

\author{Dimitris Vamvourellis}
\email{dimitrios.vamvourellis@blackrock.com}
\affiliation{%
  \institution{BlackRock, Inc.}
  \city{New York}
  \state{NY}
  \country{USA}
}

\author{Dhagash Mehta}
\email{dhagash.mehta@blackrock.com}
\affiliation{%
  \institution{BlackRock, Inc.}
  \city{New York}
  \state{NY}
  \country{USA}
}

\renewcommand{\shortauthors}{Vamvourellis \& Mehta}

\begin{abstract}
  We investigate the effectiveness of large language models (LLMs), including reasoning-based and non-reasoning models, in performing zero-shot financial sentiment analysis. Using the Financial PhraseBank dataset annotated by domain experts, we evaluate how various LLMs and prompting strategies align with human-labeled sentiment in a financial context. We compare three proprietary LLMs (GPT-4o, GPT-4.1, o3-mini) under different prompting paradigms that simulate System 1 (fast and intuitive) or System 2 (slow and deliberate) thinking and benchmark them against two smaller models (FinBERT-Prosus, FinBERT-Tone) finetuned on financial sentiment analysis. Our findings suggest that reasoning, either through prompting or inherent model design, does not improve performance on this task. Surprisingly, the most accurate and human-aligned combination of model and method was GPT-4o without any Chain-of-Thought (CoT) prompting. We further explore how performance is impacted by linguistic complexity and annotation agreement levels, uncovering that reasoning may introduce overthinking, leading to suboptimal predictions. This suggests that for financial sentiment classification, fast, intuitive "System 1"-like thinking aligns more closely with human judgment on financial sentiment analysis compared to “System-2” slower, deliberative reasoning simulated by reasoning models or CoT prompting. Our results challenge the default assumption that more reasoning always leads to better LLM decisions, particularly in high-stakes financial applications.
\end{abstract}


\maketitle

\section{Introduction}
Large language models (LLMs) have demonstrated remarkable performance in a wide range of natural language processing tasks, including question answering, summarization, and sentiment analysis \cite{brown2020gpt3, openai2023gpt4, touvron2023llama}. Much of this success has been attributed not only to scale, but also to the emergence of reasoning capabilities when prompted appropriately. In particular, Chain-of-Thought (CoT) prompting, where models are asked to generate intermediate reasoning steps before outputting a final answer, has shown notable promise for arithmetic and logical reasoning tasks \cite{wei2022cot, kojima2022zeroshot}.

Although CoT prompting has been shown to improve performance on tasks that require factual inference or logical reasoning, its impact on more subjective tasks remains less understood. One such domain is financial sentiment analysis, where the goal is not to solve a logical puzzle, but to interpret the sentiment of short financial news excerpts, mimicking intuitive judgments of human annotators. In such cases, fast and automatic thinking (System 1) may be more aligned with human labels than slow and structured reasoning processes (System 2) \cite{kahneman2011thinking}. This raises an important question: Does prompting LLMs to reason step-by-step improve alignment with human judgments, or does it introduce unnecessary complexity that harms performance on tasks grounded in perception and intuition?

Prompting strategies in LLMs increasingly reflect the architecture of dual-process theories from cognitive science: System 1 is fast, automatic, and relies on surface-level pattern matching, which corresponds to direct instruction-only prompting without intermediate reasoning. System 2 is slow, deliberative, and structured, analogous to prompting methods that explicitly invoke reasoning steps, such as CoT or self-reflective mechanisms \cite{renze2024self, shinn2023reflexion}. However, the strength of this analogy remains under investigation. Recent work \cite{binz2023cogpsych,chen2024not,sui2025stop} suggests that while CoT may help in logic-heavy domains, it can degrade performance in tasks where intuitive pattern recognition suffices. Since LLM reasoning often emerges from learned discourse patterns rather than true symbolic inference, invoking System 2-style prompts may result in overthinking or verbose rationalizations without improving predictive accuracy. This dual process lens provides a useful framework for analyzing when LLM reasoning is effective and when it becomes a liability.

In the present work, we investigate this question by evaluating the performance of several state-of-the-art LLMs on the Financial PhraseBank dataset \cite{malo2014phrasebank}, a widely used benchmark for financial sentiment classification. The dataset contains short financial sentences annotated with sentiment labels (positive, neutral, or negative) by multiple finance-savvy individuals. We focus exclusively on zero-shot performance of LLMs, to assess how well different models and prompting strategies align with human sentiment perception without being fine-tuned or primed with examples from the dataset.

We compare three proprietary OpenAI LLMs, GPT-4o, GPT-4.1, and o3-mini, under different prompting strategies that vary in reasoning structure and token output patterns. These strategies span a spectrum from immediate label prediction (mimicking System 1 inference) to extended, structured reasoning (System 2). To contextualize the LLM results, we also benchmark against two BERT-based finetuned sentiment models: FinBERT-Prosus \cite{araci2019finbert}, trained directly on the dataset used for testing, and FinBERT-Tone \cite{huang2023finbert}, finetuned on a related but out-of-domain financial corpus.

Our analysis reveals a counterintuitive but consistent pattern: prompting LLMs to engage in explicit reasoning often reduces alignment with human-labeled sentiment, especially in low-ambiguity cases. The highest agreement with human annotations is achieved by GPT-4o under No-CoT prompting, which mirrors System 1 thinking—fast, intuitive, and automatic. In contrast, models encouraged to simulate System 2–style deliberation, particularly o3-mini (which is optimized for internal reasoning), tend to produce longer completions, underperform in overall accuracy, and display signs of overthinking. This pattern suggests that invoking deliberative reasoning in contexts where intuitive judgment suffices may interfere with correct classification. Notably, the Language Model Intuitive Reasoning with Attribution (LIRA)\cite{lopez2023lira} prompting structure, where a model first commits to a judgment and then explains it, outperforms forward-chained CoT prompting in this setting. This aligns with cognitive findings that post hoc rationalization can better match human interpretability than reasoning that precedes commitment. Collectively, our findings suggest that for subjective financial tasks, simulated deliberation is not always beneficial, and in many cases, System 1–like intuition more closely tracks human consensus.

In summary, our findings challenge the default assumption that more reasoning improves model performance. On tasks grounded in human perception and intuition—like financial sentiment analysis—less can be more. Our work highlights the importance of matching prompting structure to the nature of the task, and raises broader questions about when and how LLM reasoning should be invoked. These results suggest that System 1–style prompting may be better suited for alignment tasks, whereas System 2–style reasoning should be applied judiciously, especially in domains where human intuition, not logical deduction, is the gold standard.

\section{Related Work}
In this Section, we provide a brief overview of the existing work in this area.\newline
\textbf{Chain-of-Thought Reasoning in Large Language Models}  
The success of LLMs has been driven by scale and broad pretraining \cite{brown2020gpt3,openai2023gpt4,touvron2023llama}. Prompting techniques such as Chain-of-Thought (CoT) prompting, which elicit intermediate reasoning steps, can boost performance in arithmetic, commonsense, and logic tasks \cite{wei2022cot,kojima2022zeroshot}. Extensions like Least-to-Most prompting \cite{zhou2023least2most} and Tree-of-Thought \cite{yao2023treeofthought} have been proposed to further improve accuracy by decomposing problems into explicit subgoals. Additionally, researchers have attempted to explain the effectiveness of CoT methods in a more systematic manner. Theoretical work characterizes CoT as an external computational scaffold: \cite{feng2023towards} use circuit complexity theory to show that bounded-depth Transformers benefit from CoT for arithmetic reasoning, while \cite{ton2024understanding} provide an information-theoretic interpretation that quantifies the value of each reasoning step without labeled data.

\textbf{LLMs for Financial Applications}  
LLMs have been applied to a range of financial tasks, from sentiment analysis to stock prediction. \cite{araci2019finbert} introduced FinBERT, a domain-specific model fine-tuned for financial sentiment analysis, later expanded in \cite{huang2023finbert} for extracting structured information from disclosures. BloombergGPT \cite{wu2023bloomberggpt}, trained on a large hybrid dataset, improved performance across financial benchmarks. Other domain-adapted models such as FinLlama \cite{konstantinidis2024finllama}, FinTral \cite{bhatia2024fintral}, and Fin-R1 \cite{liu2025fin} advanced applications in sentiment classification, multimodal integration, and reasoning.

Meanwhile, general-purpose models like ChatGPT have shown utility in market prediction and analysis. \cite{lopez2023lira} and \cite{glasserman2023assessing} used LLM-generated sentiment from news to predict stock returns, revealing biases like look-ahead and distraction. Social media sentiment was leveraged similarly in Refs.~\cite{mumtaz2023potential,steinert2023linking}. Beyond prediction, \cite{kim2023bloated} found that ChatGPT-generated summaries of corporate disclosures were more informative and aligned with market reactions, while \cite{kim2024context} showed that narrative context significantly enhanced the informativeness of financial numbers. Finally, \cite{fatouros2024can} proposed MarketSenseAI, a GPT-4-based framework combining CoT prompting and in-context learning to generate interpretable and profitable investment signals.

Unlike prior work focused on domain adaptation or prediction, we evaluate how reasoning depth (via prompting and model design) affects zero-shot LLM performance on financial sentiment classification, comparing intuitive versus deliberative reasoning modes.

\textbf{Reasoning versus Overthinking}  
Recent work has explored enhancing LLM reasoning via reinforcement learning and test-time compute scaling. DeepSeek-R1 \cite{guo2025deepseek} illustrates that RL alone can elicit strong reasoning behaviors. Other methods use prompt extension \cite{muennighoff2025s1}, token-budget control \cite{han2024token}, or inference strategies \cite{chen2024simple, wu2024scaling} to improve performance without retraining. These studies show that test-time compute can outperform model scaling under fixed budgets \cite{snell2024scaling}. Reasoning depth also plays a key role: longer reasoning chains boost performance in complex tasks \cite{jin2024impact}, but can introduce overthinking and inefficiency in tasks that do not require quantitative reasoning \cite{chen2024not, sui2025stop}. To address this, \cite{zhang2024othinkr1} introduce OThink-R1, a fast/slow thinking switch that dynamically classifies reasoning traces as redundant or essential, pruning unnecessary steps without harming accuracy. Their results demonstrate that LLMs can reduce reasoning length by 23\% on average while maintaining task performance. Our work adds to this literature by empirically evaluating whether deeper reasoning improves or harms performance on financial sentiment classification, a task often reliant on fast, intuitive judgments.

Before we proceed, it is important to emphasize that  \textit{while we draw comparisons between LLM prompting strategies and dual-process theories of human cognition, LLMs do not engage in genuine logical deliberation. Instead, their reasoning is simulated via language generation patterns learned from data, not necessarily via internal symbolic or deductive processes.}

\begin{figure*}[t]
  \centering
  \includegraphics[width=0.9\textwidth]{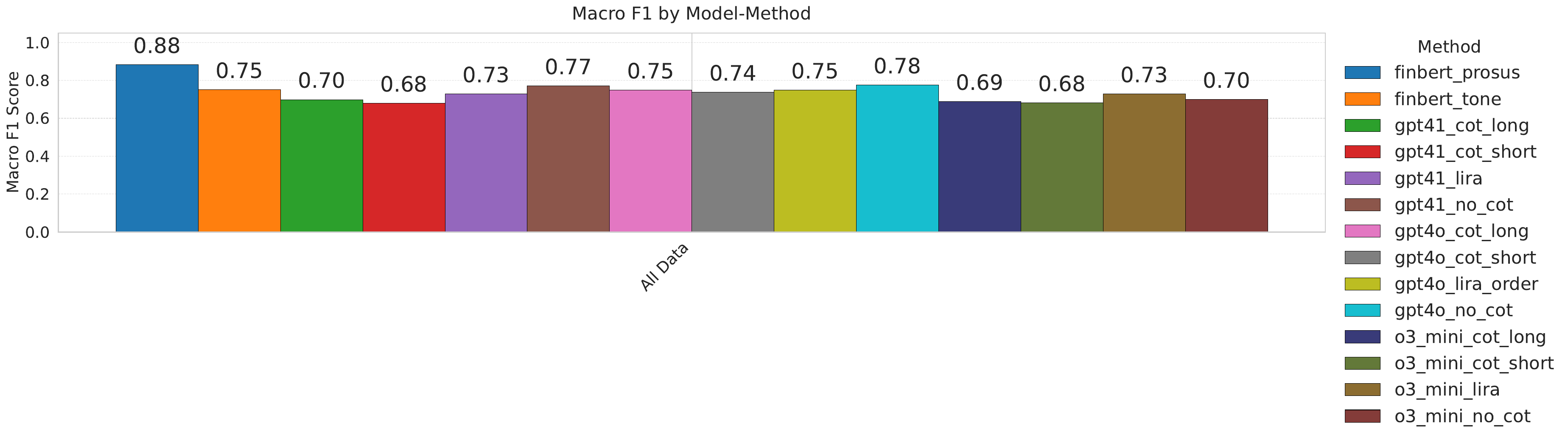}
  \caption{Macro F1 scores across models and prompting strategies over the entire Financial Phrasebank dataset.}
  \label{fig:overall}
\end{figure*}

\section{Data Description}
We use the Financial PhraseBank dataset \cite{malo2014phrasebank}, a widely used benchmark for financial sentiment analysis. It contains 4,845 English sentences extracted from LexisNexis financial news articles, each labeled positive, neutral, or negative based on the majority vote of 5–8 finance-savvy annotators.

A key feature of the dataset is the inclusion of inter-annotator agreement levels, which categorize each sentence into one of four groups: 100\%, 75–99\%, 66–74\%, and 50–65\% agreement. These strata serve as a natural proxy for sentence-level ambiguity, with lower agreement often indicating higher linguistic or contextual uncertainty.
To further analyze the relationship between sentence complexity and model performance, we compute the Flesch-Kincaid Readability Score \cite{kincaid1975derivation} (which estimates how difficult a sentence is to understand based on word and sentence length) for each sentence and divide the dataset into quartiles by increasing complexity: low, medium-low, medium-high, and high. This captures syntactic and lexical difficulty across the dataset. Together, these two dimensions allow us to investigate how performance varies for reasoning and non-reasoning models with increasing levels of ambiguity and linguistic complexity in financial text.

\begin{table}[htbp]
\centering
\begin{tabular}{>{\raggedright\arraybackslash}p{2cm}ccccc}
\toprule
\textbf{Agreement level} & \textbf{Positive} & \textbf{Negative} & \textbf{Neutral} & \textbf{Count} \\
\midrule
100\%        & \%25.2 & \%13.4 & \%61.4 & 2262 \\
75\% - 99\%  & \%26.6 & \%9.8  & \%63.6 & 1191 \\
66\% - 74\%  & \%36.7 & \%12.3 & \%50.9 & 765  \\
50\% - 65\%  & \%31.1 & \%14.4 & \%54.5 & 627  \\
\midrule
\textbf{All} & \%28.1 & \%12.4 & \%59.4 & 4845 \\
\bottomrule
\end{tabular}
\caption{Sentiment distribution by agreement level.}
\label{tab:agreement_levels}
\end{table}

\section{Methodology}
In this Section, we describe our methodology by starting with the list of models and prompting paradigms experimented within this study, then describing our experimental set-up and finally the evaluation methodology.

\subsection{Models}
We evaluate a range of language models (OpenAI snapshot for each model is provided below) spanning both proprietary LLMs and domain-specific fine-tuned smaller language models to explore differences in reasoning ability and domain generalization:
\begin{itemize}
    \item \textbf{GPT-4o} (\texttt{gpt-4o-2024-05-13}): A high-performance general-purpose model, notable for its strong zero-shot performance without explicit optimization for step-by-step reasoning.
    \item \textbf{GPT-4.1} (\texttt{gpt-4.1-2025-04-14}): A successor to GPT-4o, this model reflects the latest iteration in OpenAI's general-purpose LLM line, optimized for a broad range of tasks.
    \item \textbf{o3-mini} (\texttt{o3-mini-2025-01-31}): A lightweight LLM trained specifically for reasoning tasks, featuring inductive biases that promote internal chain-of-thought (CoT) reasoning.
    \item \textbf{FinBERT-Prosus}: A BERT-based model fine-tuned on 3,101 sentences from the Financial PhraseBank dataset. Given its exposure to much of the evaluation data during training, it serves as a benchmark rather than a direct competitor to the LLMs evaluated in a zero-shot fashion.
    \item \textbf{FinBERT-Tone}: A BERT model fine-tuned on 10,000 analyst report sentences labeled with sentiment. It provides a useful reference for understanding cross-domain generalization of task-specific fine-tuning. Specifically, we use this model to better understand how a smaller model fine-tuned to perform sentiment analysis on analyst reports can generalize to language used in news without access to the test data. Thus we are benchmarking a BERT-based model in the same way as its LLM counterparts that do not have access to the test dataset.
\end{itemize}

\subsection{Prompting Paradigms}
Our analysis adopts a zero-shot framework to isolate the intrinsic reasoning tendencies of language models, independent of external supervision. While few-shot prompting can guide model behavior with curated exemplars, it introduces additional variables such as example selection and label leakage, which may obscure our central focus: how reasoning depth and structure affect alignment in the absence of external correction. Similarly, prompt optimization techniques that explicitly leverage gold-standard annotations can improve task-specific performance, but represent a different paradigm focused on adaptation rather than evaluating emergent reasoning behavior.

We compare four prompting paradigms to analyze how prompt structure and generation length (i.e., reasoning verbosity) influence sentiment classification performance and alignment with human judgment:
\begin{itemize}
    \item \textbf{No-CoT}: The model is prompted to directly classify the sentiment without generating any intermediate reasoning steps. This simulates the behavior of standard discriminative models like BERT.
    \item \textbf{CoT-Short}: The model is prompted to generate a brief chain of reasoning before outputting a label. This approach conditions the label on a concise explanation, mirroring typical CoT prompting.
    \item \textbf{CoT-Long}: The model is encouraged to generate a more elaborate reasoning path before producing the label. This setting tests whether longer reasoning chains result in improved or degraded performance.
    \item \textbf{LIRA}: Based on Ref.~\cite{lopez2023lira}, the reverse-CoT strategy prompts the model to first generate a label and then produce an explanation. It provides a contrast to traditional CoT where the label is conditioned on prior reasoning.
\end{itemize}

The above prompting strategies can be mapped\footnote{We use cognitive analogies (e.g., System 1 and System 2) to interpret model behavior, but these should not be mistaken for literal cognitive processing.} to corresponding modes of reasoning observed in dual-process theories as shown in Table \ref{tab:system_prompt_mapping} which outlines the proposed mapping between LLM prompt types and their cognitive analogs. The prompt used for each method is presented in Appendix \ref{sec:prompt-specs}.
\begin{table}[h]
\centering
\caption{Mapping of Prompt Types to Dual-System Cognitive Analogies.}
\label{tab:system_prompt_mapping}
\begin{tabular}{|p{1cm}|p{2cm}|p{2cm}|p{2.5cm}|}
\hline
\textbf{Prompt Type} & \textbf{LLM Behavior} & \textbf{Cognitive Analog} & \textbf{Key Characteristics} \\
\hline
\textbf{No-CoT} & Direct label prediction without explicit reasoning & System 1 & Fast, automatic, heuristic-based; relies on surface patterns or priors \\
\hline
\textbf{CoT-Short} & Label prediction preceded by brief, shallow reasoning & System 1.5 (System 1 with light System 2)\tablefootnote{Note that the original dual-process theory distinguishes between System 1 (fast, intuitive) and System 2 (slow, deliberative) reasoning \citep{kahneman2011thinking, stanovich2000individual}. We use terms like “System 1.5” heuristically to denote intermediate or augmented reasoning modes simulated by LLMs, e.g., shallow reasoning steps (System 1.5) or post hoc justification and attribution (System 1+). These extensions are not formal constructs in cognitive psychology, but serve as useful descriptors in the LLM context.} & Limited decomposition; semi-structured intuition with minimal deliberation \\
\hline
\textbf{CoT-Long} & Label prediction preceded by detailed, multi-step reasoning & System 2 & Slow, logical, multi-step inferencing; simulates full deliberation \\
\hline
\textbf{LIRA} & Label predicted first, then explanation generated conditioned on prediction (often with salient spans) & System 1+ (System 1 with Metacognition\tablefootnote{Metacognition refers to the ability to monitor and regulate one’s own cognitive processes, such as evaluating confidence, detecting errors, or justifying decisions \citep{flavell1979metacognition, nelson1990metamemory}. In the context of LLMs, metacognitive prompting strategies simulate this capacity by encouraging post hoc explanations or self-attribution of reasoning. Recent work suggests that LLM-generated explanations can simulate metacognitive behavior, offering insights into internal confidence or attribution \citep{lampinen2022can}.}) & Reflective reasoning; adds post hoc rationalization, attribution, and justification after intuitive response \\
\hline
\end{tabular}
\end{table}

Using the above prompting paradigms we attempt to investigate the following research questions:
\begin{itemize}
    \item To what extent does explicit reasoning, elicited through CoT prompting, affect model performance? Does it enhance or impair alignment with human annotations?
    
    \item How does the structure of the prompt influence performance? Specifically, does it matter if the label is conditioned on the reasoning, or the reasoning on the label?
    
    \item What is the impact of longer generations, encouraged through certain prompting styles, on model performance? Do extended outputs lead to improved or degraded results?
\end{itemize}

These questions guide our systematic examination of how model behavior varies as a function of prompt structure and the relative positioning of reasoning and decision tokens.

\begin{table}[t]
  \centering
  \caption{Performance (Macro F1) of each model-method combination across annotator agreement levels.}
  \label{tab:results-by-agreement}
  \begin{tabular}{lccccc}
    \toprule
    \textbf{Model} & \textbf{Method} & \textbf{50--65} & \textbf{66--74} & \textbf{75--99} & \textbf{100} \\
    \midrule
    FinBERT-Prosus & N/A & 0.722 & 0.786 & 0.885 & 0.962 \\
    FinBERT-Tone & N/A & 0.436 & 0.566 & \textbf{0.736} & \textbf{0.897} \\
    GPT-4.1 & CoT-Long & 0.534 & 0.594 & 0.650 & 0.799 \\
    GPT-4.1 & CoT-Short & 0.504 & 0.591 & 0.635 & 0.776 \\
    GPT-4.1 & LIRA & 0.537 & 0.630 & 0.679 & 0.830 \\
    GPT-4.1 & No-CoT & 0.552 & 0.658 & 0.734 & 0.890 \\
    GPT-4o & CoT-Long & 0.505 & 0.650 & 0.718 & 0.866 \\
    GPT-4o & CoT-Short & 0.548 & 0.639 & 0.706 & 0.841 \\
    GPT-4o & LIRA & 0.560 & 0.654 & 0.689 & 0.865 \\
    GPT-4o & No-CoT & \textbf{0.561} & \textbf{0.677} & 0.727 & 0.895 \\
    o3-mini & CoT-Long & 0.506 & 0.619 & 0.621 & 0.784 \\
    o3-mini & CoT-Short & 0.482 & 0.615 & 0.624 & 0.776 \\
    o3-mini & LIRA & 0.542 & 0.636 & 0.684 & 0.825 \\
    o3-mini & No-CoT & 0.504 & 0.615 & 0.648 & 0.797 \\
    \bottomrule
  \end{tabular}
\end{table}

\begin{table}[t]
  \centering
  \small              
  \setlength{\tabcolsep}{3pt}
  \caption{Performance (Macro F1) of each model–method combination across sentence–complexity quartiles.}
  \label{tab:results-by-complexity}
  \begin{tabular}{lccccc}
    \toprule
    \textbf{Model} & \textbf{Method} & \textbf{Low} & \textbf{Medium-Low} & \textbf{Medium-High} & \textbf{High} \\
    \midrule
    FinBERT-Prosus & N/A & 0.920 & 0.886 & 0.845 & 0.870 \\
    FinBERT-Tone   & N/A & 0.829 & 0.738 & 0.685 & 0.674 \\
    GPT-4.1 & CoT-Long & 0.803 & 0.675 & 0.624 & 0.640 \\
    GPT-4.1 & CoT-Short & 0.783 & 0.660 & 0.605 & 0.620 \\
    GPT-4.1 & LIRA & 0.818 & 0.719 & 0.664 & 0.680 \\
    GPT-4.1 & No-CoT & 0.842 & 0.760 & 0.718 & 0.727 \\
    GPT-4o & CoT-Long & 0.827 & 0.720 & 0.701 & 0.700 \\
    GPT-4o & CoT-Short & 0.819 & 0.714 & 0.663 & 0.714 \\
    GPT-4o & LIRA & 0.837 & 0.728 & 0.683 & 0.710 \\
    GPT-4o & No-CoT & \textbf{0.853} & \textbf{0.768} & \textbf{0.708} & \textbf{0.732} \\
    o3-mini & CoT-Long & 0.777 & 0.678 & 0.619 & 0.642 \\
    o3-mini & CoT-Short & 0.783 & 0.674 & 0.601 & 0.630 \\
    o3-mini & LIRA & 0.812 & 0.719 & 0.663 & 0.688 \\
    o3-mini & No-CoT & 0.792 & 0.688 & 0.635 & 0.648 \\
    \bottomrule
  \end{tabular}
\end{table}

\subsection{Experimental Setup}
All LLMs are evaluated in zero-shot mode. This setup isolates the intrinsic reasoning capabilities of these models that are learned during pretraining without any biasing from in-context examples from the test dataset.

For each sentence in the dataset, we record:
\begin{itemize}
    \item The predicted sentiment label (positive, neutral, or negative);
    \item The number of completion tokens are the tokens that the model generates in response to the input prompt and are directly returned from the OpenAI API. For reasoning models, like o3-mini, this also includes the tokens generated for internal reasoning, which are not part of the final model response; and,
    \item The reasoning or explanation content (depending on the prompting strategy).
\end{itemize}

To evaluate the impact of reasoning verbosity, we stratify completion token counts into five quantile-based bins by model and prompting method. Performance is then analyzed across these bins to understand the trade-offs between reasoning verbosity and alignment with human-annotated labels.

\subsection{Evaluation}
Due to the label imbalance in the dataset (59.4\% neutral, 28.1\% positive, 12.4\% negative), we adopt macro F1 score as our primary evaluation metric. This ensures equal weight across classes, mitigating bias toward the dominant neutral class.
It is important to clarify that the Financial PhraseBank labels reflect human-perceived sentiment, not actual financial outcomes. Annotators were instructed to rate sentences based on expected short-term stock price reactions, but the dataset does not include any real market impact data.
As such, this work focuses solely on alignment with human interpretation of sentiment in financial text, not on price movement prediction. All evaluation metrics (e.g., macro-F1) measure the agreement between LLM outputs and these human-annotated sentiment labels.
Although LLMs may eventually be deployed in financial forecasting contexts, such an application would require temporally linked sentence-market pairs and backtesting under realistic trading constraints. We leave this for future work, and restrict our scope here to zero-shot sentiment classification using human-labeled financial text.

\section{Results}
In this section, we summarize the main findings of our experiments while in later sections, we focus on failure modes and sources of response variability between models and prompting strategies.\newline
\textbf{Fast thinking outperforms reasoning.} As shown in Figure \ref{fig:overall}, the best overall performance is achieved using the No-CoT prompting strategy, which yields the highest macro-F1 score among all evaluated configurations (without considering FinBERT-Prosus which has been directly trained on the same dataset). This result highlights that direct classification—akin to fast, intuitive decision-making (System 1)—aligns more closely with human-annotated sentiment labels than the more deliberative, step-by-step reasoning (System 2) simulated by Chain-of-Thought prompting strategies.

\textbf{Reasoning reduces alignment.}
Across both GPT-4o and GPT-4.1, introducing explicit reasoning via CoT-Short or CoT-Long consistently degrades performance across all agreement levels and readability categories (Tables~\ref{tab:results-by-agreement} and~\ref{tab:results-by-complexity}). Interestingly, the performance gap between CoT and No-CoT prompting methods is wider for GPT-4.1 than for GPT-4o. 

Consistent with these findings, o3-mini—optimized for internal reasoning—yields the lowest performance overall, as well as within each agreement and readability category. Notably, o3-mini continues to use internal CoT even when prompted with the No-CoT method. For this reason, its performance matches that of GPT-4.1 when the latter is prompted using CoT methods.

These results suggest that reasoning may introduce unnecessary cognitive overhead in this task, leading to misalignment with human sentiment perception—a phenomenon we refer to as \textit{overthinking}.

\textbf{Prompt structure influences performance.}
The LIRA prompting strategy, which reverses the typical CoT order by generating the label before the explanation, outperforms both CoT-Short and CoT-Long across almost every agreement level and readability category. This indicates that the reasoning on the label, rather than the reverse, is better aligned with human sentiment labels. The result reinforces the intuition that System 1-style inference, with post hoc rationalization, may more faithfully replicate human annotations. It also confirms that prompt structure and the relative positioning of reasoning and label can significantly affect the behavior and performance of LLMs due to their autoregressive nature.

\textbf{Reasoning could be helpful in higher ambiguity settings.}
As expected, model performance improves monotonically with annotator agreement levels, as shown in Table \ref{tab:results-by-agreement}. Sentences with full agreement (100\%) are easier to classify, while those with low agreement (50–65\%) introduce ambiguity. Notably, the performance gap between No-CoT and CoT prompting widens at higher agreement levels, suggesting that reasoning may be less harmful—or even modestly helpful—in ambiguous cases, but more harmful in unambiguous ones.

\textbf{Reasoning does not help in lower readability settings.}
Across all models and prompting methods, performance declines on sentences with higher linguistic complexity, as measured by the Flesch-Kincaid readability score (Table~\ref{tab:results-by-complexity}). Notably, reasoning—whether explicit or implicit through a reasoning-optimized model—does not confer improved robustness to increased syntactic or semantic complexity in this setting.

\begin{figure}[t]
  \centering
  \includegraphics[width=\linewidth]{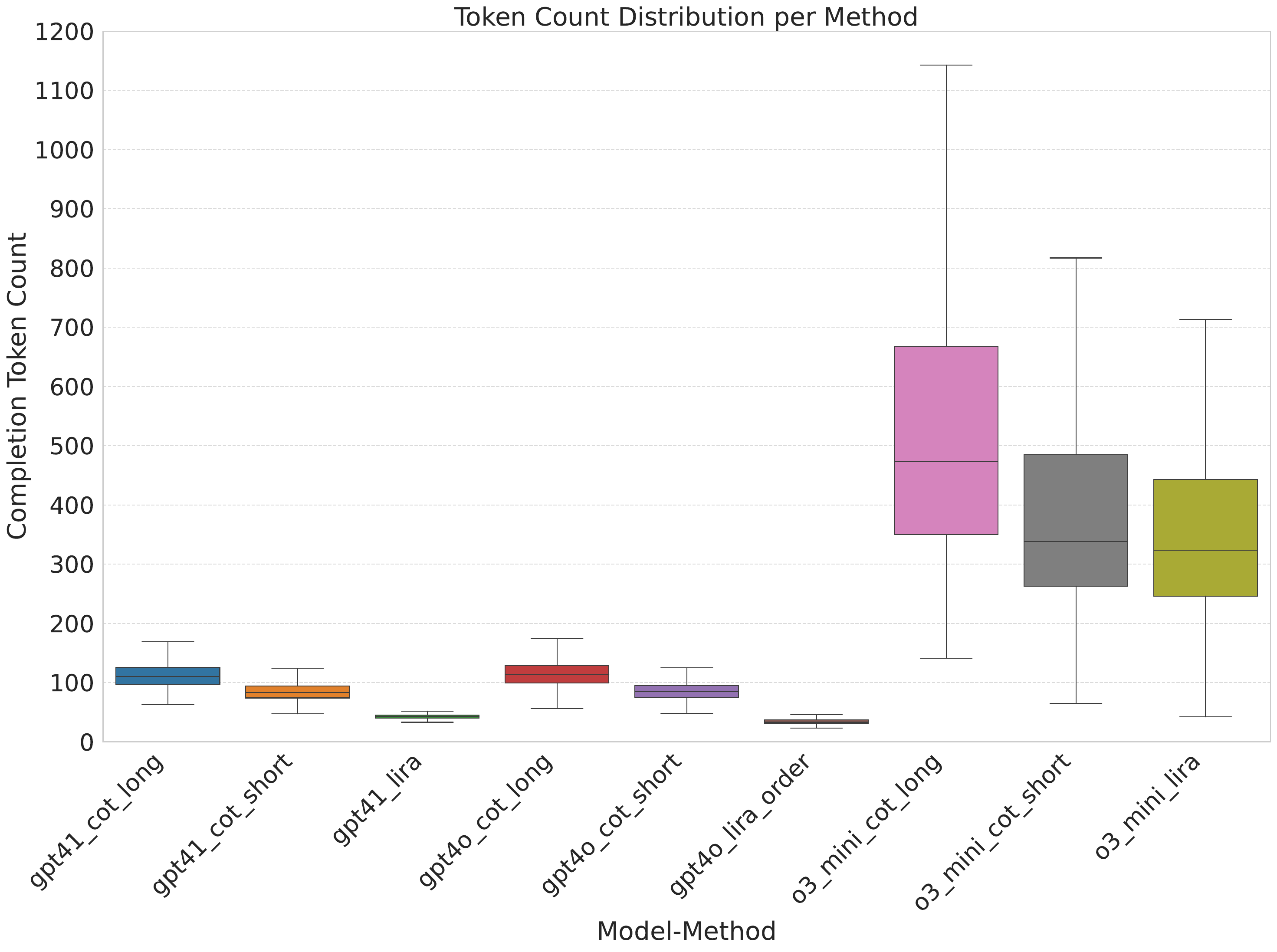}
  \caption{Macro F1 scores across models and prompting strategies. Results show significant variation depending on the combination of model and method, underscoring the importance of aligning the choice of approach with task-specific goals.}
  \label{fig:token-boxplot}
\end{figure}

\textbf{Longer reasoning chains hurt human-alignment.}
Performance monotonically declines with increasing completion length across all models and prompting strategies (Table \ref{tab:results-by-token-q}). The number of tokens generated is negatively correlated with macro-F1, reinforcing the notion that longer, more verbose reasoning may reflect overthinking rather than deeper understanding.
As shown on Figure \ref{fig:token-boxplot}, o3-mini, the most verbose model, generates 4–5 times more tokens on average compared to GPT-4o and GPT-4.1 across all prompting strategies, yet achieves the lowest performance.
Among prompting strategies, CoT-Long consistently produces more tokens than CoT-Short, and both exceed LIRA in output length across all LLMs used. This pattern further supports the hypothesis that increased reasoning depth may dilute model alignment with human sentiment interpretation and lead to overthinking in tasks that are more intuitive, such as sentiment analysis.

\textbf{Finetuning helps in-domain, however LLMs demonstrate better generalization.} FinBERT-Prosus achieves the highest macro-F1 overall, however it has been trained on a substantial portion of the test set. As such, it serves as a reference point rather than a fair competitor to the LLMs prompted in zero-shot fashion without access to the test set. FinBERT-Tone, trained on sentiment-labeled analyst reports, is competitive on less ambiguous cases with high agreement levels between human annotators or in cases with lower linguistic complexity, however it is significantly worse than GPT-4o and GPT-4.1 with No-CoT prompting in more complex or ambiguous cases (category 50-65 and 66-74 in Table \ref{tab:results-by-agreement} and High complexity category in Table \ref{tab:results-by-complexity}). This suggests that task-specific finetuning may not transfer well across domains, and that general-purpose LLMs, even in zero-shot mode, can outperform traditional finetuned smaller models on out-of-domain inputs with higher linguistic complexity or semantic ambiguity.

\begin{table}[t]
  \centering
  \small              
  \setlength{\tabcolsep}{5pt}
  \caption{Macro-F1 performance of each model–method combination across completion-token-count quantile bins (Q1 = fewest tokens, Q5 = most).}
  \label{tab:results-by-token-q}
  \begin{tabular}{l l ccccc}
    \toprule
    \textbf{Model} & \textbf{Method} & \textbf{Q1} & \textbf{Q2} & \textbf{Q3} & \textbf{Q4} & \textbf{Q5} \\
    \midrule
    FinBERT-Prosus & N/A        & 0.883 & N/A   & N/A   & N/A   & N/A   \\
    FinBERT-Tone & N/A          & 0.751 & N/A   & N/A   & N/A   & N/A   \\
    GPT-4.1 & CoT-Long      & 0.820 & 0.775 & 0.708 & 0.592 & 0.490 \\
    GPT-4.1 & CoT-Short     & 0.761 & 0.712 & 0.707 & 0.602 & 0.482 \\
    GPT-4.1 & LIRA          & 0.773 & 0.721 & 0.730 & 0.700 & 0.697 \\
    GPT-4.1 & No-CoT        & 0.772 & N/A   & N/A   & N/A   & N/A   \\
    GPT-4o  & CoT-Long      & 0.812 & 0.799 & 0.737 & 0.678 & 0.619 \\
    GPT-4o  & CoT-Short     & 0.757 & 0.760 & 0.726 & 0.743 & 0.631 \\
    GPT-4o  & LIRA (Order)  & 0.704 & 0.725 & 0.773 & 0.749 & 0.754 \\
    GPT-4o  & No-CoT        & 0.776 & N/A   & N/A   & N/A   & N/A   \\
    o3-mini & CoT-Long      & 0.850 & 0.763 & 0.673 & 0.540 & 0.446 \\
    o3-mini & CoT-Short     & 0.827 & 0.744 & 0.663 & 0.576 & 0.429 \\
    o3-mini & LIRA          & 0.858 & 0.801 & 0.723 & 0.627 & 0.499 \\
    o3-mini & No-CoT        & 0.802 & 0.732 & 0.640 & 0.531   & 0.464   \\
    \bottomrule
  \end{tabular}
\end{table}

\begin{table}[t]
  \centering
  \caption{Confusion matrix for \textbf{GPT-4o \& No-CoT} predictions (values in \%).}
  \label{tab:conf-matrix-gpt4o-nocot}
  \begin{tabular}{lcccc}
    \toprule
            & \textbf{Pred.\ Pos} & \textbf{Pred.\ Neg}
            & \textbf{Pred.\ Neu} & \textbf{Total} \\
    \midrule
    Actual positive & 22.7 & 0.1 & 5.3 & 28.2 \\
    Actual negative & 0.1 & 11.8 & 0.6 & 12.5 \\
    Actual neutral  & 11.8 & 4.7 & 42.9 & 59.3 \\
    \midrule
    \textbf{Column total} & 34.6 & 16.5 & 48.8 & 100.0 \\
    \bottomrule
  \end{tabular}
\end{table}

\begin{table}[t]
  \centering
  \caption{Confusion matrix for \textbf{GPT-4o \& CoT-Short} predictions (values in \%).}
  \label{tab:conf-matrix-gpt4o-cot-short}
  \begin{tabular}{lcccc}
    \toprule
            & \textbf{Pred.\ Pos} & \textbf{Pred.\ Neg} & \textbf{Pred.\ Neu} & \textbf{Total} \\
    \midrule
    Actual positive & 22.6 & 0.1 & 5.4 & 28.2 \\
    Actual negative & 0.1  & 11.0 & 1.4 & 12.5 \\
    Actual neutral  & 13.1 & 5.8 & 40.5 & 59.3 \\
    \midrule
    \textbf{Column total} & 35.8 & 16.9 & 47.3 & 100.0 \\
    \bottomrule
  \end{tabular}
\end{table}

\begin{table}[t]
  \centering
  \caption{Confusion matrix for \textbf{GPT-4.1 \& No-CoT} predictions (values in \%).}
  \label{tab:conf-matrix-gpt41-nocot}
  \begin{tabular}{lcccc}
    \toprule
            & \textbf{Pred.\ Pos} & \textbf{Pred.\ Neg} & \textbf{Pred.\ Neu} & \textbf{Total} \\
    \midrule
    Actual positive & 24.1 & 0.2 & 3.9 & 28.2 \\
    Actual negative & 0.2  & 11.4 & 0.8 & 12.5 \\
    Actual neutral  & 14.4 & 4.0 & 41.0 & 59.3 \\
    \midrule
    \textbf{Column total} & 38.7 & 15.6 & 45.7 & 100.0 \\
    \bottomrule
  \end{tabular}
\end{table}

\begin{table}[t]
  \centering
  \caption{Confusion matrix for \textbf{GPT-4.1 \& CoT-Short} predictions (values in \%).}
  \label{tab:conf-matrix-gpt41-cot-short}
  \begin{tabular}{lcccc}
    \toprule
            & \textbf{Pred.\ Pos} & \textbf{Pred.\ Neg} & \textbf{Pred.\ Neu} & \textbf{Total} \\
    \midrule
    Actual positive & 25.5 & 0.6 & 2.0 & 28.2 \\
    Actual negative & 0.7  & 11.3 & 0.5 & 12.5 \\
    Actual neutral  & 21.1 & 7.8 & 30.4 & 59.3 \\
    \midrule
    \textbf{Column total} & 47.4 & 19.7 & 33.0 & 100.0 \\
    \bottomrule
  \end{tabular}
\end{table}

\begin{table}[t]
  \centering
  \caption{Confusion matrix for \textbf{o3-mini \& No-CoT} predictions (values in \%).}
  \label{tab:conf-matrix-o3mini-nocot}
  \begin{tabular}{lcccc}
    \toprule
            & \textbf{Pred.\ Pos} & \textbf{Pred.\ Neg} & \textbf{Pred.\ Neu} & \textbf{Total} \\
    \midrule
    Actual positive & 26.8 & 0.1 & 1.3 & 28.2 \\
    Actual negative & 0.4  & 11.7 & 0.4 & 12.5 \\
    Actual neutral  & 25.3 & 5.1 & 28.9 & 59.3 \\
    \midrule
    \textbf{Column total} & 52.5 & 17.0 & 30.6 & 100.0 \\
    \bottomrule
  \end{tabular}
\end{table}

\begin{table}[t]
  \centering
  \caption{Confusion matrix for \textbf{o3-mini \& CoT-Short} predictions (values in \%).}
  \label{tab:conf-matrix-o3mini-cot-short}
  \begin{tabular}{lcccc}
    \toprule
            & \textbf{Pred.\ Pos} & \textbf{Pred.\ Neg} & \textbf{Pred.\ Neu} & \textbf{Total} \\
    \midrule
    Actual positive & 26.7 & 0.1 & 1.3 & 28.2 \\
    Actual negative & 0.3  & 11.8 & 0.4 & 12.5 \\
    Actual neutral  & 26.1 & 6.1 & 27.1 & 59.3 \\
    \midrule
    \textbf{Column total} & 53.2 & 18.0 & 28.8 & 100.0 \\
    \bottomrule
  \end{tabular}
\end{table}

\section{Analysis and Discussion}
In this section, we provide our analysis on the above results and discuss their implications.
\subsection{Confusion Matrix Insights}
To better understand model behavior and the nature of classification errors, we analyze confusion matrices across all evaluated models and prompting strategies. We present sample confusion matrices for each model under the No-CoT and CoT-Short methods in Tables~\ref{tab:conf-matrix-gpt4o-nocot}--\ref{tab:conf-matrix-o3mini-cot-short}, while the remaining confusion matrices are provided in Appendix \ref{sec:app-confmatrices}.

The most frequent misclassification across all configurations is the erroneous prediction of \textit{positive} sentiment in place of the correct \textit{neutral} label. This pattern aligns with the characteristics of the data set and observed inter-annotator disagreements: the agreement rates for distinguishing \textit{positive-negative}, \textit{negative-neutral}, and \textit{positive-neutral} are 98.7\%, 94.2\%, and 75.2\%, respectively. The relative ambiguity between \textit{positive} and \textit{neutral} labels reflects a challenge noted by annotators themselves, namely, distinguishing routine optimistic language (e.g., ``company glitter'') from genuinely positive sentiment.

CoT-style prompts do not significantly bias GPT-4o in one direction compared to No-CoT methods. Specifically, GPT-4o misclassifies neutral cases as positive in approximately 11.8--14\% of instances, depending on the prompting strategy. The o3-mini model, by contrast, shows a significantly stronger bias toward \textit{positive} predictions across all prompting strategies, misclassifying neutral cases as positive in 22--26\% of instances. Further investigation is needed to determine whether this systematic bias stems from training data composition, inductive biases introduced during pretraining, or specific architectural choices.

Interestingly, GPT-4.1 mirrors GPT-4o under No-CoT prompting, misclassifying neutral cases as positive about 14\% of the time. However, its behavior shifts significantly under CoT or LIRA prompts, becoming more aligned with o3-mini and misclassifying neutral sentences as positive about 19--22\% of the time. This suggests that GPT-4.1 may have been exposed to reasoning traces from models like o3-mini during supervised fine-tuning, or that it shares architectural characteristics that influence its reasoning behavior. Thus, when prompted to reason before responding, it may behave more like o3-type models than GPT-4o.

\subsection{Failure Modes}

To deepen our understanding of model-specific behaviors and how prompt structure influences prediction outcomes, we analyze five representative failure cases drawn from our dataset. These examples, presented in Appendix \ref{sec:app-failures}, highlight critical patterns in misclassification and response variability across prompting methods and model choice:
\begin{itemize}
   \item \textbf{Prompt-induced variability.} Example 1 in Table \ref{tab:failure-cases} involves a sentence with 100\% inter-annotator agreement on a \textit{positive} label. Yet, GPT-4o produced three different predictions across four prompting strategies: No-CoT predicted \textit{positive}, CoT-Short and LIRA predicted \textit{negative}, and CoT-Long predicted \textit{neutral}. This variability—despite model and input remaining constant—highlights the strong influence of prompt structure on model output. Such discrepancies underscore the need for careful prompt selection based on task-specific goals and success criteria. Just as hyperparameters are tuned to optimize objective metrics in classical machine learning, prompt strategies for LLMs must be chosen in a systematic way based on clear success criteria.

   \item \textbf{Reasoning in subjective tasks.} In Examples 2 and 5, we observe high variability across models and prompting strategies in interpreting modest financial headlines, such as those reporting a small increase in sales. While human annotators generally label even small increases (e.g., 1\%) as \textit{positive}, models with longer reasoning chains sometimes label these as \textit{neutral} or even \textit{negative}, reasoning that such minor gains may lack significance—particularly if not driven by organic growth, which is an economically sound conclusion. This illustrates that longer chain-of-thought reasoning may not necessarily be incorrect—its validity could only be assessed with access to real financial outcomes—but it is nonetheless be less aligned with the fast, heuristic judgments that humans apply in subjective tasks. Therefore, the choice of model and prompting method should be guided by the intended use case and evaluation criteria—including whether alignment with human judgment is a primary objective.

   \item \textbf{o3-mini positivity bias.} In  Example 3, most models and prompting strategies correctly predicted a \textit{neutral} label, yet o3-mini predicted \textit{positive} under all prompting methods. Its explanation adopted an overly expansive interpretation, inferring that the company’s mere presence in a few U.S. towns signaled business growth. This case illustrates a systematic bias in o3-mini toward the \textit{positive} class and highlights how extended reasoning chains can lead to \textit{overthinking}: rather than anchoring the prediction to the textual content, the model extrapolates beyond what is stated, thereby misclassifying a factually neutral statement based on generated assumptions.

   \item \textbf{FinBERT fails to do the math.}
   Lastly, we document a consistent failure mode in FinBERT which struggles to interpret the compositional meaning of financial phrases especially in cases where there is cognitive dissonance between negative wording and positive outcome. (Example 4). Particularly, FinBERT systematically misclassified cases that involve improvements expressed through negatively framed terms in the corporate language, like ``smaller loss'' or ``falling costs.'' This failure appears to stem from an overreliance on isolated lexical cues (``loss'', ``costs'', ``smaller''), rather than an ability to understand how combinations of negative words can modify sentiment in a positive way. In contrast, all LLMs correctly handled such cases, demonstrating an ability to reason over word combinations and understand their financial implications, even without being fine-tuned on this dataset. This highlights the advantage of LLMs in capturing deeper semantic relationships and performing lightweight reasoning in contexts where surface-level cues may be misleading.

\end{itemize}

Taken together, these examples reinforce the need to treat prompting and model selection as critical design choices—ones that should be systematically validated and tailored to each use-case, especially in subjective tasks like financial sentiment analysis.

\section{Conclusion}
In this work, we evaluate a range of proprietary and domain-specific models on the task of financial sentiment classification, with a particular focus on the role of reasoning. Contrary to trends observed in other NLP tasks, our findings indicate that \textit{more} reasoning—whether explicitly prompted via chain-of-thought (CoT) methods or implicitly encoded through reasoning-optimized model architectures—is less aligned with human-annotated sentiment labels.

Models employing direct, fast classification (e.g., GPT-4o with No-CoT prompting) consistently outperform their reasoning-intensive counterparts. This suggests that System 1-style intuitive inference—fast, automatic, and heuristic-driven—more effectively mirrors the strategies used by human annotators when interpreting sentiment in financial texts. In contrast, System 2-style deliberate reasoning, introduced via chain-of-thought prompting or reasoning-optimized models, often injects noise or uncertainty, ultimately degrading both classification accuracy and alignment with human judgments—even for sentences with perfect annotator agreement.

These findings caution against the default use of chain-of-thought prompting and reasoning models for all tasks. While such approaches may enhance performance in domains requiring complex logical deduction or multi-step computation, they are not universally effective. Our results emphasize that the decision to invoke reasoning, the structure of prompt and even the relative positioning of reasoning tokens within the prompt should be treated as systematic design choices, evaluated using objective metrics that reflect the specific goals of each task.

\section*{Acknowledgment}
The views expressed here are those of the authors alone and not of BlackRock, Inc.
\bibliographystyle{ACM-Reference-Format}
\bibliography{main}
\appendix

\onecolumn
\section{Prompt Specifications}\label{sec:prompt-specs}

\begingroup
\scriptsize
\setlength{\tabcolsep}{4pt}        
\begin{tabularx}{\linewidth}{l>{\raggedright\arraybackslash}X>{\raggedright\arraybackslash}X}
\toprule
\textbf{Method} & \textbf{System Prompt} & \textbf{User Prompt}\\
\midrule
\textbf{LIRA} &
Forget all your previous instructions. Pretend you are a financial expert. You are a financial expert with stock-recommendation experience. Answer ``YES'' if good news, ``NO'' if bad news, or ``UNKNOWN'' if uncertain in the first line. Then, elaborate with 1--2 concise sentences on the next line.\newline
Respond only with a JSON object that has two attributes: \texttt{Response} (values: \texttt{YES}, \texttt{NO}, \texttt{UNKNOWN}) and \texttt{Explanation} (the reasoning). &
Is this headline good or bad for the stock price of \emph{company name} in the short term?\newline
Headline: <\texttt{headline}>\\[4pt]

\textbf{CoT-Short} &
Forget all your previous instructions. Pretend you are a financial expert with stock-recommendation experience. You are going to be presented with a headline about a company.\newline
First, think step by step for a \textbf{short} amount of time and provide some reasoning on whether this headline is good or bad for the stock price of the company in the short term.\newline
Then, on a new line, provide a final answer which is ``YES'' if good news, ``NO'' if bad news, or ``UNKNOWN'' if uncertain.\newline
Respond only with a JSON object that has two attributes: \texttt{Reasoning} and \texttt{Response}. &
Is this headline good or bad for the stock price of \emph{company name} in the short term?\newline
Headline: <\texttt{headline}>\\[4pt]

\textbf{CoT-Long} &
Forget all your previous instructions. Pretend you are a financial expert with stock-recommendation experience. You are going to be presented with a headline about a company.\newline
First, think step by step for a \textbf{long} amount of time and provide some reasoning on whether this headline is good or bad for the stock price of the company in the short term.\newline
Then, on a new line, provide a final answer which is ``YES'' if good news, ``NO'' if bad news, or ``UNKNOWN'' if uncertain.\newline
Respond only with a JSON object that has two attributes: \texttt{Reasoning} and \texttt{Response}. &
Is this headline good or bad for the stock price of \emph{company name} in the short term?\newline
Headline: <\texttt{headline}>\\[4pt]

\textbf{No-CoT} &
Forget all your previous instructions. Pretend you are a financial expert with stock-recommendation experience. You are going to be presented with a headline about a company.\newline
Provide a final answer which is ``YES'' if this headline is good for the stock price of the company in the short term, ``NO'' if it is bad, or ``UNKNOWN'' if uncertain.\newline
Respond only with a JSON object that has one attribute: \texttt{Response}. &
Is this headline good or bad for the stock price of \emph{company name} in the short term?\newline
Headline: <\texttt{headline}>\\[4pt]
\bottomrule
\end{tabularx}
\endgroup

\section{Model-Prompt Induced Response Variability}\label{sec:app-failures}
\begingroup
\scriptsize
\setlength{\tabcolsep}{2pt}
\setlength{\LTcapwidth}{\linewidth}
\centering
\begin{longtable}{p{5cm} p{1.2cm} p{2.6cm} p{8.0cm} p{0.75cm}}

\caption{Model responses (predicted label and explanation) generated from each model and prompting method for sample sentences from Financial Phrasebank dataset. The number of completion tokens are any tokens that the model generates in response to the input prompt and they are directly returned from the OpenAI API. For reasoning models, like o3-mini, this also includes the tokens generated for internal reasoning, which are not part of the final model response provided in this table.}\label{tab:failure-cases}\\   
\toprule
\textbf{Original Sentence} & \textbf{Label} & \textbf{Model--Method} & \textbf{Model Response} & \textbf{\#Tokens}\\
\midrule
\endfirsthead
\multicolumn{5}{l}{\textbf{Table \thetable{} — continued from previous page}}\\
\toprule
\textbf{Original Sentence} & \textbf{Label} & \textbf{Model--Method} & \textbf{Model Response} & \textbf{\#Tokens}\\
\midrule
\endhead
\midrule
\multicolumn{5}{r}{\textit{Continued on next page}}\\
\midrule
\endfoot
\bottomrule
\endlastfoot

\multirow{13}{5cm}{\textbf{Example 1:} In January , traffic , measured in revenue passenger kilometres RPK , went up by 3.2\% and capacity , measured in available seat kilometres ASK , rose by 12.2\%.} & \multirow{13}{1.5cm}{\textbf{Positive}} & FinBERT-Prosus & \textbf{Positive} &  \\
 &  & FinBERT-Tone & \textbf{Positive} &  \\
 &  & GPT-4.1 \& CoT-Long & \textbf{Negative} - The headline indicates that revenue passenger kilometres (RPK), a measure of demand, increased by 3.2\%, while available seat kilometres (ASK), a measure of capacity, rose by 12.2\%. This means that capacity is growing much faster than demand, which could lead to lower load factors (percentage of seats filled) and potentially lower pricing power or yields. In the short term, this imbalance is generally viewed negatively by investors, as it may signal overcapacity and pressure on profitability. & 115 \\
 &  & GPT-4.1 \& CoT-Short & \textbf{Negative} - The headline indicates that while traffic (RPK) increased by 3.2\%, capacity (ASK) rose much faster at 12.2\%. This suggests that the company added more capacity than demand grew, which could lead to lower load factors and potential pressure on ticket prices and margins. Investors may view this as a negative in the short term. & 86 \\
  &  & GPT-4.1 \& LIRA & Negative - Capacity increased much faster than traffic, which may indicate lower load factors and potential pressure on short-term profitability. & 36 \\
 &  & GPT-4.1 \& No-CoT & \textbf{Negative} & 7 \\
 &  & GPT-4o \& CoT-Long & \textbf{Neutral} - The headline indicates that the company's traffic, measured in revenue passenger kilometers (RPK), increased by 3.2\% in January. This is a positive sign as it suggests higher demand for the company's services. However, the capacity, measured in available seat kilometers (ASK), rose by 12.2\%, which is a significantly higher increase compared to the traffic growth. This could imply that the company has increased its capacity more than the demand, potentially leading to lower load factors and reduced efficiency. In the short term, this mixed news could be interpreted as uncertain because while the increase in traffic is positive, the disproportionate increase in capacity might raise concerns about overcapacity and its impact on profitability. & 155 \\
 &  & GPT-4o \& CoT-Short & \textbf{Negative} - The headline indicates that the company's traffic, measured in revenue passenger kilometers (RPK), increased by 3.2\%, which is positive as it suggests higher demand for the company's services. However, the capacity, measured in available seat kilometers (ASK), rose by 12.2\%, which is a much higher increase compared to the traffic growth. This could imply that the company is expanding its capacity faster than the demand is growing, potentially leading to lower load factors and reduced profitability in the short term. & 116 \\
 &  & GPT-4o \& LIRA & \textbf{Negative} - Capacity growth outpaced traffic growth, which may indicate lower utilization and potential inefficiencies. & 32 \\
 &  & GPT-4o \& No-CoT & \textbf{Positive} & 10 \\
 &  & o3-mini \& CoT-Long & \textbf{Negative} The headline shows that while traffic (indicative of demand) increased moderately by 3.2\%, capacity expanded much more significantly at 12.2\%. In the airline industry, this discrepancy may lead to a reduction in load factor—the percentage of seats filled—which is an important efficiency and profitability metric. In the short term, an oversupply of seats relative to demand could negatively impact revenue per available seat and investor sentiment. Therefore, the news is likely to be viewed as negative in the short term. & 574 \\
 &  & o3-mini \& CoT-Short & \textbf{Negative} - Although the headline shows an increase in both passenger traffic (RPK up 3.2\%) and available capacity (ASK up 12.2\%), the disproportionate growth in capacity suggests that the load factor is declining. In the short term, this may indicate an inefficient use of capacity, potentially pressuring yields and profitability. Investors could view the modest increase in traffic relative to the significant capacity expansion as a negative sign. & 1006 \\
 &  & o3-mini \& LIRA & \textbf{Negative} - Despite an increase in capacity, the much smaller rise in passenger traffic could indicate declining load factors, which is generally a short-term negative for profitability. & 502 \\
 &  & o3-mini \& No-CoT & \textbf{Neutral} & 721 \\
\midrule
\multirow{13}{5cm}{\textbf{Example 2:} Net sales went up by 1\% year-on-year to EUR 29 million , affected by \dots} & \multirow{13}{1.5cm}{\textbf{Positive}} & FinBERT-Prosus & \textbf{Positive} &  \\
 &  & FinBERT-Tone & \textbf{Positive} &  \\
 &  & GPT-4.1 \& CoT-Long & \textbf{Positive} - A 1\% year-on-year increase in net sales is a positive sign, indicating growth. However, the growth is modest and largely attributed to business acquisitions from the previous period, which contributed EUR 5.1 million to the current period's sales. This suggests that organic growth may be limited, and the headline does not mention profitability or other financial metrics. Investors may view the reliance on acquisitions for growth as a potential concern. Overall, the news is slightly positive but not strongly so. & 114 \\
 &  & GPT-4.1 \& CoT-Short & \textbf{Negative} - A 1\% year-on-year increase in net sales is a modest growth, which may not impress investors, especially if the increase is largely attributed to business acquisitions rather than organic growth. The headline suggests that the underlying business may not be growing significantly on its own. This could be viewed as neutral to slightly negative in the short term, as investors often look for strong organic growth. & 92 \\
 &  & GPT-4.1 \& LIRA & \textbf{Positive} - A year-on-year increase in net sales, even if modest and partly due to acquisitions, is generally positive for short-term stock sentiment. & 42 \\
 &  & GPT-4.1 \& No-CoT & \textbf{Positive} & 7 \\
 &  & GPT-4o \& CoT-Long & \textbf{Neutral} - The headline indicates that the company's net sales increased by 1\% year-on-year to EUR 29 million. This increase is partly attributed to business acquisitions made in the previous financial period, which contributed EUR 5.1 million to the review period. While an increase in net sales is generally positive, the fact that a significant portion of this increase is due to acquisitions rather than organic growth might be a concern for some investors. Additionally, a 1\% increase is relatively modest and may not be seen as a strong indicator of growth. Therefore, the impact on the stock price in the short term is uncertain. & 138 \\
 &  & GPT-4o \& CoT-Short & \textbf{Positive} - The headline indicates that net sales have increased by 1\% year-on-year, which is generally a positive indicator for the company's performance. However, the increase is partly attributed to business acquisitions, which contributed EUR 5.1 million to the net sales. This suggests that the organic growth might be lower than the headline figure implies. While the increase in sales is good news, the reliance on acquisitions for growth could be seen as a mixed signal by investors. Overall, the headline leans towards being positive but with some caution. & 121 \\
 &  & GPT-4o \& LIRA & \textbf{Positive} - An increase in net sales, even by a small percentage, is generally positive for the stock price. & 35 \\
 &  & GPT-4o \& No-CoT & \textbf{Positive} & 10 \\
 &  & o3-mini \& CoT-Long & \textbf{Negative} - The headline shows that while traffic (indicative of demand) increased moderately by 3.2\%, capacity expanded much more significantly at 12.2\%. In the airline industry, this discrepancy may lead to a reduction in load factor—the percentage of seats filled—which is an important efficiency and profitability metric. In the short term, an oversupply of seats relative to demand could negatively impact revenue per available seat and investor sentiment. Therefore, the news is likely to be viewed as negative in the short term. & 574 \\
 &  & o3-mini \& CoT-Short & \textbf{Negative} - Although the headline shows an increase in both passenger traffic (RPK up 3.2\%) and available capacity (ASK up 12.2\%), the disproportionate growth in capacity suggests that the load factor is declining. In the short term, this may indicate an inefficient use of capacity, potentially pressuring yields and profitability. Investors could view the modest increase in traffic relative to the significant capacity expansion as a negative sign. & 1006 \\
 &  & o3-mini \& LIRA & \textbf{Neutral} - The net sales grew modestly by 1\%, but the growth is driven by prior acquisitions rather than organic performance, leaving uncertainty about short-term momentum. & 692 \\
 &  & o3-mini \& No-CoT & \textbf{Negative} & 529 \\
\midrule
\multirow{13}{5cm}{\textbf{Example 3:} Los Angeles-based Pacific Office Properties Trust acquires , owns , and operates office properties in Honolulu , San Diego , Los Angeles , and Phoenix .} & \multirow{13}{1.5cm}{\textbf{Neutral}} & FinBERT-Prosus & \textbf{Neutral} &  \\
 &  & FinBERT-Tone & \textbf{Neutral} &  \\
 &  & GPT-4.1 \& CoT-Long & \textbf{Neutral} - The headline is a general description of Pacific Office Properties Trust's business model and geographic focus. It does not mention any new developments, acquisitions, financial results, or other news that would directly impact the stock price in the short term. Therefore, there is no clear positive or negative implication for the stock price based on this information alone. & 82 \\
 &  & GPT-4.1 \& CoT-Short & \textbf{Neutral} - The headline is descriptive and provides information about the company's business model and geographic focus, but it does not mention any new developments, financial results, acquisitions, or other news that would directly impact the stock price in the short term. Therefore, it is not possible to determine if this is good or bad news for the stock price based on this headline alone. & 86 \\
 &  & GPT-4.1 \& LIRA & \textbf{Neutral} - The headline is descriptive and does not indicate any new developments or events that would impact the stock price in the short term. & 39 \\
 &  & GPT-4.1 \& No-CoT & \textbf{Neutral} & 7 \\
 &  & GPT-4o \& CoT-Long & \textbf{Positive} - The headline indicates that Pacific Office Properties Trust is expanding its portfolio by acquiring, owning, and operating office properties in key markets in the western U.S. This expansion can be seen as a positive move for the company as it may lead to increased revenue and market presence. Investors often view acquisitions and expansions as a sign of growth and potential profitability, which can drive the stock price up in the short term. Additionally, focusing on major markets like Honolulu, San Diego, Los Angeles, and Phoenix suggests strategic planning and targeting areas with potentially high demand for office spaces. & 127 \\
 &  & GPT-4o \& CoT-Short & \textbf{Neutral} - The headline indicates that Pacific Office Properties Trust is expanding its portfolio by acquiring, owning, and operating office properties in key markets in the western U.S. This expansion could be seen as a positive move, as it may lead to increased revenue and market presence. However, the headline does not provide specific details about the financial impact or the terms of the acquisitions, making it difficult to assess the immediate effect on the stock price. Therefore, the impact on the stock price in the short term is uncertain. & 115 \\
 &  & GPT-4o \& LIRA & \textbf{Neutral} - The headline provides information about the company's operations but does not indicate any recent developments or news that would impact the stock price in the short term. & 43 \\
 &  & GPT-4o \& No-CoT & \textbf{Neutral} & 10 \\
 &  & o3-mini \& CoT-Long & \textbf{Positive} - The headline announces that Pacific Office Properties Trust is actively acquiring and operating office properties in key Western U.S. markets, including regions like Los Angeles and Phoenix which are known for their economic dynamism. Such expansion efforts generally signal growth and strategic positioning, potentially leading to enhanced revenues and market share. In the short term, investors often view acquisition news positively as it reflects the company's proactive approach in capitalizing on market opportunities. & 557 \\
 &  & o3-mini \& CoT-Short & \textbf{Positive} - The headline indicates that Pacific Office Properties Trust is actively expanding its portfolio through acquisitions and operations across strategically important western U.S. markets. Such expansion efforts are generally viewed as positive in the short term because they signal growth and increased market presence, which can boost investor sentiment. & 719 \\
 &  & o3-mini \& LIRA & \textbf{Positive} - The acquisition of office properties in key markets suggests growth and expansion, which tends to boost investor confidence in the short term. & 625 \\
 &  & o3-mini \& No-CoT & \textbf{Positive} & 529 \\
\midrule
\multirow{13}{5cm}{\textbf{Example 4:} The loss for the third quarter of 2007 was EUR 0.3 mn smaller than the loss of the second quarter of 2007 .} & \multirow{13}{1.5cm}{\textbf{Positive}} & FinBERT-Prosus & \textbf{Negative} &  \\
 &  & FinBERT-Tone & \textbf{Negative} &  \\
 &  & GPT-4.1 \& CoT-Long & \textbf{Positive} - The headline indicates that the company's loss in the third quarter of 2007 was smaller than in the previous quarter, suggesting an improvement in financial performance. While the company is still operating at a loss, the reduction in losses can be interpreted positively by investors as a sign of potential turnaround or improving operational efficiency. In the short term, this could lead to a positive reaction in the stock price, as it may increase investor confidence. & 101 \\
 &  & GPT-4.1 \& CoT-Short & \textbf{Positive} - The headline indicates that the company's loss has decreased compared to the previous quarter, which suggests an improvement in the company's financial performance. While the company is still operating at a loss, the reduction in losses may be viewed positively by investors as a sign of progress toward profitability. This could have a mildly positive effect on the stock price in the short term. & 83 \\
 &  & GPT-4.1 \& LIRA & \textbf{Positive} - A smaller loss compared to the previous quarter indicates improvement, which is generally viewed positively by investors in the short term. & 38 \\
 &  & GPT-4.1 \& No-CoT & \textbf{Positive} & 7 \\
 &  & GPT-4o \& CoT-Long & \textbf{Positive} - The headline indicates that the company has reduced its losses in the third quarter compared to the second quarter. This is a positive development as it shows that the company is moving towards better financial health. Investors may view this as a sign of improvement and potential future profitability, which could lead to a positive impact on the stock price in the short term. & 84 \\
 &  & GPT-4o \& CoT-Short & \textbf{Positive} - The headline indicates that the company's loss has decreased in the third quarter compared to the second quarter. This suggests an improvement in the company's financial performance, which could be seen positively by investors. However, the company is still experiencing a loss, which may temper the positive impact. & 70 \\
 &  & GPT-4o \& LIRA & \textbf{Positive}Positive - A smaller loss compared to the previous quarter indicates improvement, which is generally positive for stock prices. & 34 \\
 &  & GPT-4o \& No-CoT & \textbf{Positive} & 10 \\
 &  & o3-mini \& CoT-Long & \textbf{Positive} - The headline demonstrates that the company's quarterly loss decreased by EUR 0.3 million from the second quarter to the third quarter. This improvement, despite still being a loss, is a positive sign as it indicates that the company's financial performance is improving in the short term, potentially restoring investor confidence. & 405 \\
 &  & o3-mini \& CoT-Short & \textbf{Positive} - The headline indicates that the company's loss narrowed in the third quarter compared to the second quarter, suggesting an improvement in performance despite still incurring a loss. In the short term, such a trend can be viewed as a positive sign, potentially boosting investor sentiment. & 333 \\
 &  & o3-mini \& LIRA & \textbf{Positive} - The smaller loss compared to the previous quarter indicates an improvement in performance, which is generally seen as positive by investors in the short term. & 372 \\
 &  & o3-mini \& No-CoT & \textbf{Positive} & 81 \\
\midrule
\multirow{13}{5cm}{\textbf{Example 5:} Net sales totaled EUR 93.6 mn , up from EUR 93.2 mn in the corresponding period in 2005 .} & \multirow{13}{1.5cm}{\textbf{Positive}} & FinBERT-Prosus & \textbf{Positive} &  \\
 &  & FinBERT-Tone & \textbf{Positive} &  \\
 &  & GPT-4.1 \& CoT-Long & \textbf{Negative} - The headline reports a slight increase in net sales from EUR 93.2 million to EUR 93.6 million compared to the same period in the previous year. While this is technically growth, the increase is very marginal (approximately 0.4\%). Investors typically look for stronger growth rates, and such a small improvement may be seen as underwhelming or indicative of stagnation. In the short term, this could be interpreted as neutral to slightly negative, as it does not signal strong momentum or significant positive change. & 120 \\
 &  & GPT-4.1 \& CoT-Short & \textbf{Positive} - The headline reports a slight increase in net sales compared to the previous period. While the growth is minimal (EUR 93.6 mn vs. EUR 93.2 mn), it is still positive and indicates stability or slight improvement in revenue. However, the increase is not significant, so the positive impact on the stock price may be limited. & 85 \\
 &  & GPT-4.1 \& LIRA & \textbf{Positive} - The headline indicates a year-over-year increase in net sales, which is generally positive for the stock price in the short term. & 40 \\
 &  & GPT-4.1 \& No-CoT & \textbf{Positive} & 7 \\
 &  & GPT-4o \& CoT-Long & \textbf{Neutral} - The headline indicates a slight increase in net sales from EUR 93.2 million to EUR 93.6 million. While this is a positive development, the increase is marginal and may not significantly impact investor sentiment in the short term. Investors typically look for substantial growth or other positive indicators to drive stock prices up. Therefore, the impact on the stock price is likely to be minimal. & 93 \\
 &  & GPT-4o \& CoT-Short & \textbf{Neutral} - The headline indicates a slight increase in net sales compared to the corresponding period in 2005. While the increase is minimal, it is still positive news as it shows growth. However, the small magnitude of the increase may not significantly impact the stock price in the short term. & 71 \\
 &  & GPT-4o \& LIRA & \textbf{Neutral} - The increase in net sales is minimal, making it unclear if it will significantly impact the stock price in the short term. & 39 \\
 &  & GPT-4o \& No-CoT & \textbf{Neutral} & 10 \\
 &  & o3-mini \& CoT-Long & \textbf{Neutral}Neutral - The headline indicates that net sales increased marginally from EUR 93.2 mn to EUR 93.6 mn, which is only about a 0.4\% increase. Although any sales growth is technically positive, the improvement is minimal and may not be enough to excite investors or drive a significant short‐term stock price movement. Given that the change is very slight, there is uncertainty about whether this would be interpreted as strong performance or just a neutral outcome. & 951 \\
 &  & o3-mini \& CoT-Short & \textbf{Neutral} - The headline shows a marginal increase in net sales (EUR 93.6 mn vs EUR 93.2 mn), which indicates a slight improvement but is so minimal that it suggests nearly flat performance over the corresponding period. Such minor growth is not likely to drive significant positive sentiment in the short term, and because there's little context about margins, costs, or other financial metrics, the overall impact on the stock price remains unclear. & 495 \\
 &  & o3-mini \& LIRA & \textbf{Neutral} - The increase in net sales is minimal, which makes it unclear whether this slight improvement alone will have a significant short-term impact on the stock price. & 502 \\
 &  & o3-mini \& No-CoT & \textbf{Positive} & 465 \\
\bottomrule
\end{longtable}
\endgroup

\twocolumn

\section{Confusion Matrices}\label{sec:app-confmatrices}
Below we present the confusion matrices for every model–method combination
whose matrix was not included in the main text.



\begin{table}[H]
  \centering
  \caption{Confusion matrix for \textbf{FinBERT-Prosus} predictions (values in \%).}
  \label{tab:conf-matrix-finbert_prosus}
  \begin{tabular}{lcccc}
    \toprule
            & \textbf{Pred.\ Pos} & \textbf{Pred.\ Neg} & \textbf{Pred.\ Neu} & \textbf{Total} \\
    \midrule
    Actual positive & 25.9 & 0.5 & 1.8 & 28.2 \\
    Actual negative & 0.1 & 12.1 & 0.2 & 12.5 \\
    Actual neutral  & 5.9 & 2.5 & 50.9 & 59.3 \\
    \midrule
    \textbf{Column total} & 32.0 & 15.1 & 52.9 & 100.0 \\
    \bottomrule
  \end{tabular}
\end{table}

\begin{table}[H]
  \centering
  \caption{Confusion matrix for \textbf{FinBERT-Tone} predictions (values in \%).}
  \label{tab:conf-matrix-finbert_tone}
  \begin{tabular}{lcccc}
    \toprule
            & \textbf{Pred.\ Pos} & \textbf{Pred.\ Neg} & \textbf{Pred.\ Neu} & \textbf{Total} \\
    \midrule
    Actual positive & 22.8 & 0.4 & 5.0 & 28.2 \\
    Actual negative & 0.2 & 11.6 & 0.7 & 12.5 \\
    Actual neutral  & 11.2 & 3.7 & 44.4 & 59.3 \\
    \midrule
    \textbf{Column total} & 34.2 & 15.7 & 50.1 & 100.0 \\
    \bottomrule
  \end{tabular}
\end{table}

\begin{table}[H]
  \centering
  \caption{Confusion matrix for \textbf{GPT-4o \& CoT-Long} predictions (values in \%).}
  \label{tab:conf-matrix-gpt4o_cot_long}
  \begin{tabular}{lcccc}
    \toprule
            & \textbf{Pred.\ Pos} & \textbf{Pred.\ Neg} & \textbf{Pred.\ Neu} & \textbf{Total} \\
    \midrule
    Actual positive & 15.5 & 1.0 & 11.7 & 28.2 \\
    Actual negative & 0.6 & 8.6 & 3.4 & 12.5 \\
    Actual neutral  & 11.3 & 4.6 & 43.4 & 59.3 \\
    \midrule
    \textbf{Column total} & 27.4 & 14.2 & 58.4 & 100.0 \\
    \bottomrule
  \end{tabular}
\end{table}

\begin{table}[H]
  \centering
  \caption{Confusion matrix for \textbf{GPT-4o \& LIRA} predictions (values in \%).}
  \label{tab:conf-matrix-gpt4o_lira_order}
  \begin{tabular}{lcccc}
    \toprule
            & \textbf{Pred.\ Pos} & \textbf{Pred.\ Neg} & \textbf{Pred.\ Neu} & \textbf{Total} \\
    \midrule
    Actual positive & 18.6 & 0.5 & 9.1 & 28.2 \\
    Actual negative & 0.3 & 9.7 & 2.5 & 12.5 \\
    Actual neutral  & 13.4 & 4.4 & 41.4 & 59.3 \\
    \midrule
    \textbf{Column total} & 32.3 & 14.6 & 53.1 & 100.0 \\
    \bottomrule
  \end{tabular}
\end{table}

\begin{table}[H]
  \centering
  \caption{Confusion matrix for \textbf{GPT-4.1 \& CoT-Long} predictions (values in \%).}
  \label{tab:conf-matrix-gpt41_cot_long}
  \begin{tabular}{lcccc}
    \toprule
            & \textbf{Pred.\ Pos} & \textbf{Pred.\ Neg} & \textbf{Pred.\ Neu} & \textbf{Total} \\
    \midrule
    Actual positive & 24.1 & 0.2 & 3.9 & 28.2 \\
    Actual negative & 0.2 & 11.4 & 0.8 & 12.5 \\
    Actual neutral  & 14.4 & 4.0 & 41.0 & 59.3 \\
    \midrule
    \textbf{Column total} & 38.7 & 15.6 & 45.7 & 100.0 \\
    \bottomrule
  \end{tabular}
\end{table}

\begin{table}[H]
  \centering
  \caption{Confusion matrix for \textbf{GPT-4.1 \& LIRA} predictions (values in \%).}
  \label{tab:conf-matrix-gpt41_lira}
  \begin{tabular}{lcccc}
    \toprule
            & \textbf{Pred.\ Pos} & \textbf{Pred.\ Neg} & \textbf{Pred.\ Neu} & \textbf{Total} \\
    \midrule
    Actual positive & 25.3 & 0.3 & 2.6 & 28.2 \\
    Actual negative & 0.1 & 11.8 & 0.5 & 12.5 \\
    Actual neutral  & 15.7 & 4.1 & 39.5 & 59.3 \\
    \midrule
    \textbf{Column total} & 41.1 & 16.2 & 42.7 & 100.0 \\
    \bottomrule
  \end{tabular}
\end{table}

\begin{table}[H]
  \centering
  \caption{Confusion matrix for \textbf{o3-mini \& CoT-Long} predictions (values in \%).}
  \label{tab:conf-matrix-o3_mini_cot_long}
  \begin{tabular}{lcccc}
    \toprule
            & \textbf{Pred.\ Pos} & \textbf{Pred.\ Neg} & \textbf{Pred.\ Neu} & \textbf{Total} \\
    \midrule
    Actual positive & 24.4 & 0.6 & 3.1 & 28.2 \\
    Actual negative & 0.3 & 10.9 & 1.2 & 12.5 \\
    Actual neutral  & 17.7 & 4.9 & 36.7 & 59.3 \\
    \midrule
    \textbf{Column total} & 42.4 & 16.4 & 41.2 & 100.0 \\
    \bottomrule
  \end{tabular}
\end{table}

\begin{table}[H]
  \centering
  \caption{Confusion matrix for \textbf{o3-mini \& LIRA} predictions (values in \%).}
  \label{tab:conf-matrix-o3_mini_lira}
  \begin{tabular}{lcccc}
    \toprule
            & \textbf{Pred.\ Pos} & \textbf{Pred.\ Neg} & \textbf{Pred.\ Neu} & \textbf{Total} \\
    \midrule
    Actual positive & 23.7 & 0.5 & 4.0 & 28.2 \\
    Actual negative & 0.3 & 11.1 & 1.2 & 12.5 \\
    Actual neutral  & 16.0 & 4.3 & 39.0 & 59.3 \\
    \midrule
    \textbf{Column total} & 40.0 & 15.9 & 44.1 & 100.0 \\
    \bottomrule
  \end{tabular}
\end{table}


\end{document}